\documentclass[conference]{IEEEtran}
\IEEEoverridecommandlockouts
\usepackage{cite}
\usepackage{amsmath,amssymb,amsfonts}
\usepackage{algorithmic}
\usepackage{graphicx}
\usepackage{textcomp}
\usepackage{xcolor}

\usepackage{multirow}
\usepackage{hyperref}
\usepackage{algorithm}
\usepackage{algorithmic}

\makeatletter
\newcommand{\linebreakand}{%
  \end{@IEEEauthorhalign}
  \hfill\mbox{}\par
  \mbox{}\hfill\begin{@IEEEauthorhalign}
}
\makeatother

\def\BibTeX{{\rm B\kern-.05em{\sc i\kern-.025em b}\kern-.08em
    T\kern-.1667em\lower.7ex\hbox{E}\kern-.125emX}}
\begin{document}

\title{Distance-Aware eXplanation Based Learning}

\author{\IEEEauthorblockN{1\textsuperscript{st} Misgina Tsighe Hagos\IEEEauthorrefmark{1}} School of Computer Science \and
\IEEEauthorblockN{2\textsuperscript{nd} Niamh Belton\IEEEauthorrefmark{2}} School of Medicine \and \IEEEauthorblockN{3\textsuperscript{rd} Kathleen M. Curran\IEEEauthorrefmark{3}} School of Medicine  
\and
\IEEEauthorblockN{4\textsuperscript{th} Brian Mac Namee\IEEEauthorrefmark{4}} School of Computer Science \linebreakand \linebreakand
\IEEEauthorblockA{Science Foundation Ireland Centre for Research Training in Machine Learning \\
University College Dublin\\
Dublin, Ireland\\
Email: \IEEEauthorrefmark{1}misgina.hagos@ucdconnect.ie,
\IEEEauthorrefmark{2}niamh.belton@ucdconnect.ie,
\IEEEauthorrefmark{3}kathleen.curran@ucd.ie,
\IEEEauthorrefmark{4}brian.macnamee@ucd.ie}}

\maketitle

\begin{abstract}
eXplanation Based Learning (XBL) is an interactive learning approach that provides a transparent method of training deep learning models by interacting with their explanations. XBL augments loss functions to penalize a model based on deviation of its explanations from user annotation of image features. The literature on XBL mostly depends on the intersection of visual model explanations and image feature annotations. We present a method to add a distance-aware explanation loss to categorical losses that trains a learner to focus on important regions of a training dataset. Distance is an appropriate approach for calculating explanation loss since visual model explanations such as Gradient-weighted Class Activation Mapping (Grad-CAMs) are not strictly bounded as annotations and their intersections may not provide complete information on the deviation of a model's focus from relevant image regions. In addition to assessing our model using existing metrics, we propose an interpretability metric for evaluating visual feature-attribution based model explanations that is more informative of the model’s performance than existing metrics. We demonstrate performance of our proposed method on three image classification tasks.
\end{abstract}

\begin{IEEEkeywords}
eXplanation Based Learning, Interactive Machine Learning, eXplainable AI
\end{IEEEkeywords}

\section{Introduction}
\label{section:introduction}

Research on model transparency in deep learning is dominated by studies on dataset bias \cite{gebru2022excerpt}, model interpretability, and explainability \cite{goebel2018explainable}. Another field of study that aims to improve model transparency, Interactive Machine Learning (IML), hits two birds with one stone \cite{fails2003interactive,fiebrink2011human}. First, it provides transparency through engagement by allowing user interaction in the model training process. Second, it improves model performance by collecting expert knowledge directly from users. IML usually considers users as \emph{dumb partners} with the sole responsibility of categorizing training instances into one of a set of pre-selected categories as opposed to \emph{clever partners} who can clarify their feedback in addition to categorizing instances. However, advances in model explanation research opens the door for a more detailed and richer interaction between models and users during training.

\subsection{eXplanation based learning}

While model explanation methods have been proposed and continue to be used to tackle the ``\textit{black-box}'' nature of deep learning models \cite{singh2020explainable,van2022explainable}, they can also be used in an interactive learning approach to promote a more transparent model training process \cite{stumpf2009interacting,kulesza2015principles}. This is known as eXplanation Based Learning (XBL)\footnote{Different terms such as \emph{explanatory debugging} \cite{kulesza2015principles}, \emph{explanatory interactive learning}  \cite{teso2019explanatory}, \emph{explanatory guided learning} \cite{popordanoska2020machine} are used in the literature. We choose to use the term  \emph{eXplanation Based Learning} because we believe it generalizes all of them.}, which collects user feedback on model explanations and uses the feedback to train, debug, or refine a trained model.


In applications such as medical image classifications, deep learning models have been observed to focus on non-relevant or confounding parts of medical images such as artifacts for their classification or prediction outputs \cite{degrave2021ai,pfeuffer2023explanatory}. In addition to promoting transparent learning process, XBL has the potential to unlearn such wrong correlations, which are termed as confounding regions, confounders, or spurious correlations (used interchangeably in this paper) \cite{hagos2022identifying,pfeuffer2023explanatory}; confounding regions are parts of training instances that are not correlated with a category, but incorrectly assumed to be so by a learner.

\begin{figure*}[t]
  \centerline{\includegraphics[width=0.65\linewidth]{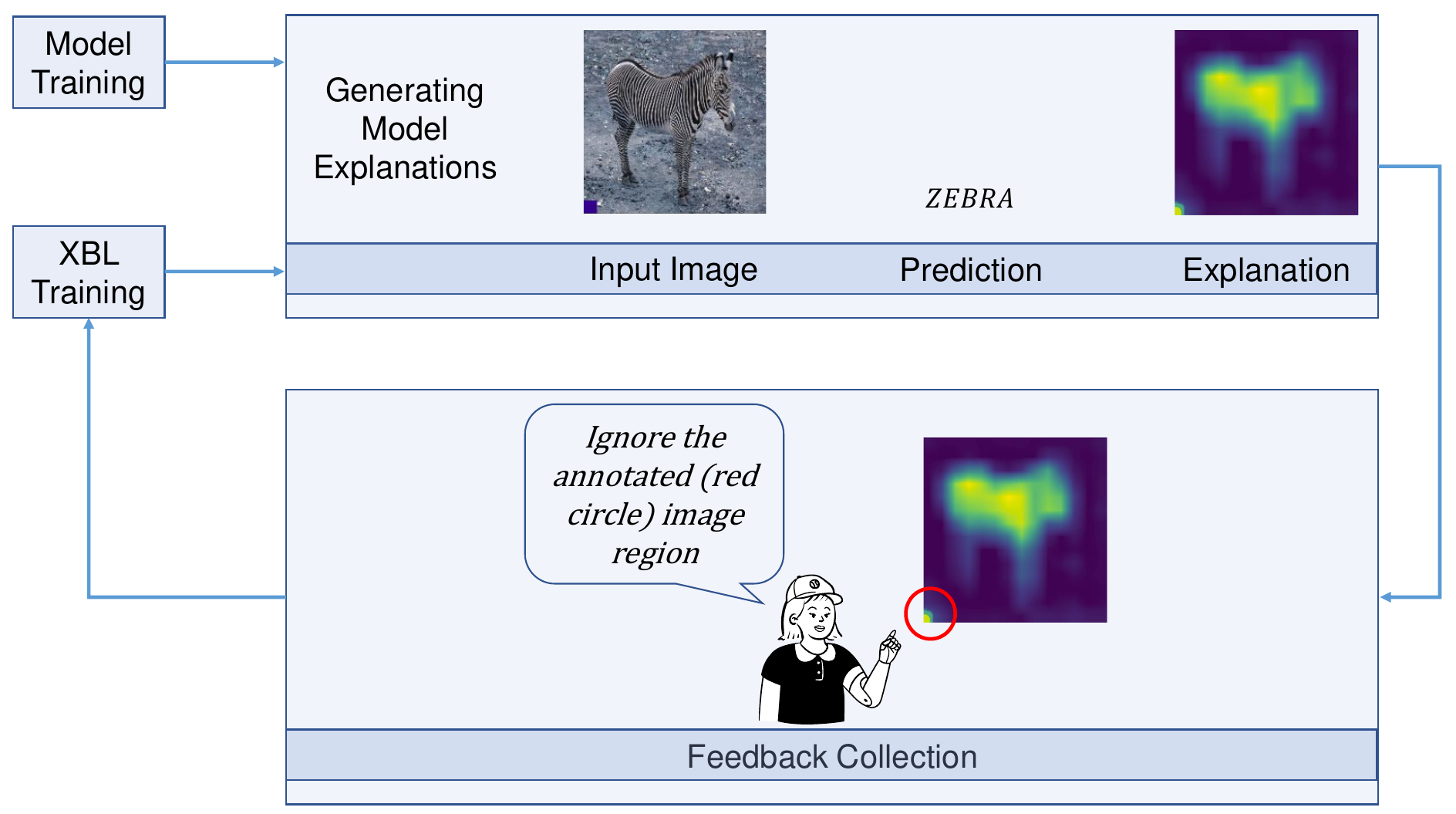}}
  \caption{The eXplanation Based Learning (XBL) loop. The user feedback, which is expected to be an annotation mask of the confounding image region in the lower left corner (highlighted by the saliency map), is portrayed here as a red circle for easier visualization.}
  \label{figure:XBLloop}
\end{figure*}

As is displayed in Fig. \ref{figure:XBLloop}, XBL is generally made up of four steps. The first is traditional model training which often uses a categorical loss. The next step is generating model explanations. Feature attribution based local explanations \cite{teso2019explanatory} or surrogate model explanations \cite{popordanoska2020machine} can be used for this. We limit the scope of this work to a saliency based local explanation, Gradient-weighted Class Activation Mapping (Grad-CAM) \cite{selvaraju2017grad}. In the third step, explanations are presented to users and feedback is collected. For method development and experiment purposes confounding regions can be added to a dataset and their masks used as user feedback for XBL. Finally, the collected feedback is used to calculate an explanation loss, which in turn is used to augment the initial categorical loss, and refine the original model using XBL training \cite{stumpf2009interacting}.

The training process in XBL augments loss functions to include an explanation loss, which can be based on either or both of: (1) a model's deviation from user annotated feedback that shows objects of interest; and (2) a model's focus on user annotation of non-salient or confounding image regions. This explanation loss is usually based on the intersection of the user annotation of image features and a model's visual explanation. Loss functions are generally augmented as follows:

\begin{eqnarray}
\label{formula:explanation_loss_generic}
    L_{expl} = \sum_{i=1}^N e(expl_{i,c} , M_{i,c}) \\
\label{categorical}
    L_{CE} = - \sum_{i=1}^N e(\hat{y}_{i}, Y_{i}) \\
\label{formula:generic_xbl}
    L = L_{expl} + L_{CE} + \lambda{}\sum_{i=1}\theta_i
\end{eqnarray}

The term, \begin{math} L_{expl} \end{math} in \eqref{formula:explanation_loss_generic} is the explanation loss  calculated as the error, $e$, between the model's explanation, \begin{math} expl_{i,c} \end{math} for input \begin{math}i\end{math} with category $c$, and the ground truth annotation, $M_{i,c}$, where $M=1$ for relevant regions and $M=0$ for non-salient regions. The term, \begin{math} L_{CE} \end{math}, in \eqref{categorical} is the traditional cross entropy loss which is calculated based on the error, $e$, between the model's prediction $\hat{y}_{i}$ and ground-truth label $Y_{i}$ for instance $i$. While $Y_{i}$ only holds category label, $M_{i,c}$ holds a mask annotation of relevant objects in an input \begin{math}i\end{math}. Finally, XBL consists of the sum of \begin{math} L_{expl} \end{math}, \begin{math} L_{CE} \end{math} and a regularization term, $\lambda$  where $\theta_{i}$ is the network parameters.

Most XBL loss function augmentations in the literature  fail to consider two scenarios: (1) focus of a model's attention may get closer to- and gradually shift to the relevant regions of training instances; for this reason, we need to penalize the learner less as the explanations (the model's attention) starts to improve. This means there is a need to make loss functions positively related to the distance of a model's wrong attention from the relevant regions; and (2) model activations that usually make up model explanations are not as strictly bounded as user annotations and we need to relax the training penalization as we get closer to the relevant parts of training images. In order to address these shortcomings of existing XBL methods and assuming model explanations correctly highlight the reasoning behind a model's output, in this paper, we address the following research question: ``\textit{Can we augment XBL loss functions in way that is sensitive to distances between explanations and user annotations of relevant image regions for better classification and explanation performance?}''

Another aspect of XBL that is often overlooked is using coefficients that weigh and balance impact of explanation losses and classification losses and optimizing them. We also consider these coefficients as hyper-parameters and tune them to find their optimal values before starting model training with XBL.

\subsection{Evaluation of model explanations}

While subjective evaluations of explanations that involve humans would give a user-centric assessment of model generated explanations \cite{halliwell2020trustworthy}, objective evaluations are often used for a speedy assessment and comparison in the development of model explainability methods \cite{dabkowski2017real}. Most of the existing evaluation methods give weight to the generated explanations over the ground truth feature annotations. This can result in over-confident evaluation results. In addition to using an existing evaluation method, to address this issue we propose an interpretability metric that assesses how much of the ground truth feature annotation has been identified as relevant by model explanations. We restrict our work in this paper to objective evaluations.


The main contributions of this paper are:
\begin{enumerate}
    \item Decoyed versions of image classification datasets are created for XBL experiments. 
    \item A new XBL method,  Distance-Aware eXplanation Based Learning (XBL-D), is proposed and evaluated.
    \item A saliency map explanation interpretability metric, Activation Recall, is proposed and demonstrated.
    \item Our experiments demonstrate that incorporating distance-aware learning into XBL performs better than baseline algorithms in classification tasks, and generates more accurate explanations. Furthermore, Code and links to download the datasets are shared online\footnote{\url{https://github.com/Msgun/XBL-D}}.
\end{enumerate}

\section{Related work}
\label{section:related_work}

In this section, we present a review of relevant literature on XBL and model explanation evaluation metrics.
\subsection{eXplanation based learning}

XBL methods can be generally categorized into two categories: (1) augmenting loss functions with explanation losses; and (2) using user feedback to augment training datasets by removing confounding or spurious regions identified by users.

\subsubsection{Augmenting loss functions}

The model explanation method used has a huge impact on an interactive learning process, not only because it is directly used to compute explanation loss ($expl_{i,c}$ as in \eqref{formula:explanation_loss_generic}), but also because it can impact user experience and feedback quality. Right for the Right Reasons (RRR) \cite{ross2017right} penalises a model with high input gradient model explanations on the wrong image regions annotated by a user. RRR uses

\begin{eqnarray}
    L_{expl} = \sum_{n}^N (M_n \frac{\partial} {\partial x_n}(\sum_{k=1}^K \log{\hat{y}_{nk}))^2}
\end{eqnarray}

\noindent for a function $f(X|\theta)=\hat{y} \in {R}^{N\times K}$ trained on images $x_{n}$ of size $N$ with $K$ categories, where $M_{n} \in \ \{0,\ 1\}$ is user annotation of image regions that should be avoided by the model.

A  Grad-CAM model explanation was used instead of input gradients in RRR-G by Schramowski \textit{et al.} \cite{schramowski2020making} using the following loss function:

\begin{eqnarray}
    L_{expl} = \sum_{n}^N M_{n}GradCAM(x_{n})
\end{eqnarray}

Similarly, Right for Better Reasons (RBR) \cite{shao2021right} uses Influence Functions (IF) in place of input gradients to correct a model's behavior. Contextual Decomposition Explanation Penalization (CDEP) \cite{rieger2020interpretations} penalizes features and feature interactions.

User feedback in XBL experiments can be one or both of: (1) telling the model to ignore non-salient image regions; and (2) instructing the model to focus on important image regions in a training dataset \cite{hagos2022impact}. While the XBL methods presented above refine a model by using the first feedback type, Human Importance-aware Network Tuning (HINT) does the opposite by teaching a model to focus on important image parts using Grad-CAM model explanations \cite{selvaraju2019taking}.

Most of the literature on XBL focuses on using feature attribution based saliency maps such as input gradients and Grad-CAMs as model explanations. Prototype based explanations have also been utilized in Bontempelli \textit{et al.} \cite{bontempelli2023concept} to debug Part-Prototype networks at concept level.

\subsubsection{Augmenting training dataset}

Instead of augmenting loss functions, XBL can be implemented by relabeling, augmenting existing instances, or adding new  training instances based on user feedback. Instance relabeling has been deployed to clean label noise in a training dataset that is identified using example based explanations \cite{teso2021interactive}. Counter-Examples (CE), which are variants of training instances with added modifications using user feedback can be generated to augment dataset for model re-training \cite{teso2019explanatory}. Simpler surrogate models have also been used as global explanations to elicit feedback in the form of new training instances \cite{popordanoska2020machine}.

\subsection{Evaluating feature attribution based explanations}

Feature attribution based explanations can be evaluated intrinsically and/or extrinsically \cite{gupta2022new}. Intrinsic evaluation involves only the model and the generated explanations themselves \cite{adebayo2018sanity}, while extrinsic evaluation involves subjective human evaluation \cite{holzinger2020measuring} or objective usage of ground-truth annotation data.

Objective evaluation of model explanations provides an easier and quicker way of assessing interpretability by comparing model explanations to ground-truth data. Overlap of visual explanations and feature annotations can be used to compute localization ability of a model's explanations; to avoid explanations with high false positive rates which cover wide area of an image, thereby scoring a high overlap with annotations, penalized versions of overlap: Penalized Localization Accuracy (PLA) was proposed \cite{belton2021optimising}. Activation Precision (AP) is another approach that computes how many of the pixels predicted as relevant by a model are actually relevant \cite{barnett2021case}. AP is presented in \eqref{formula:activation_precision}, where $A_{obj_{n}}$ is a mask of relevant image regions in input image $x_{n}$ and $T_{r}$ is a threshold function that finds the (100-$r$) percentile and sets elements of the explanation, $expl_\theta$, below this value to zero and the remaining elements to one. AP usually requires a low $r$ value or high threshold so we can avoid explanations with high false positive rates.

\begin{eqnarray}
\label{formula:activation_precision}
    AP = \frac{1}{N} \sum_{n}^N\frac{ T_{r}(expl_{\theta}(x_{n})) * A_{obj_{n}}}{T_{r}(expl_{\theta}(x_{n}))} 
\end{eqnarray} 

There is a trade-off between selecting a higher threshold and accurately assessing model explanations. While increasing the threshold would mean focusing on smaller areas of an explanation and avoiding high false positive rates, it also means parts of an explanation would be masked before they are assessed, which could result in overconfidence in model explanations.

\section{Distance-aware explanation based learning}

\begin{figure}[h!]
  \centerline{\includegraphics[width=\linewidth]{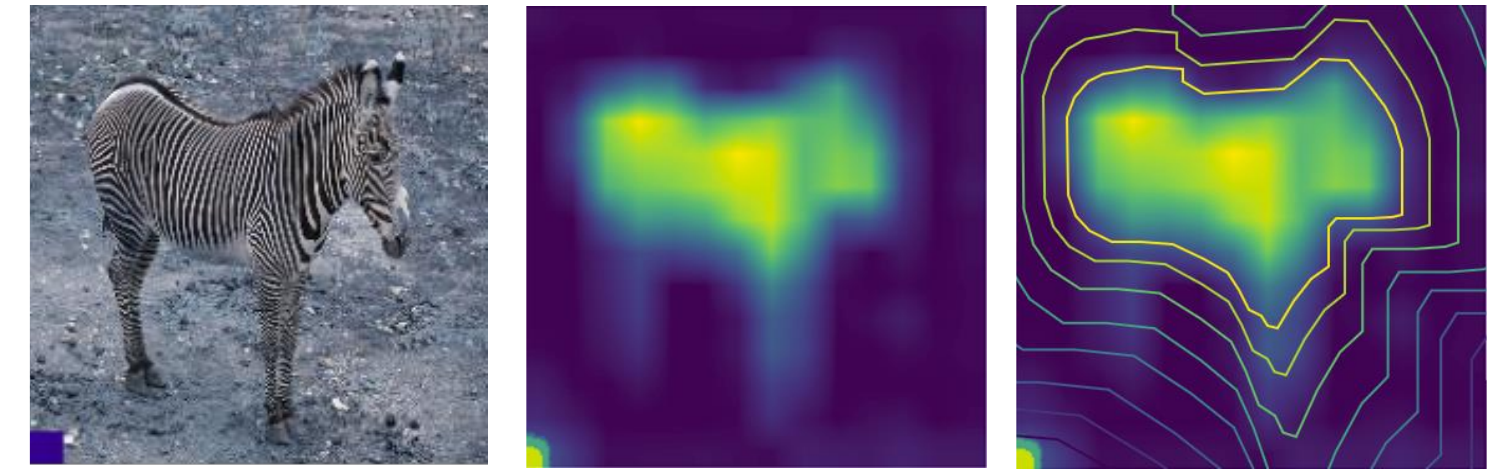}}
  \caption{[Best viewed in color.] Illustration of a distance-aware explanation loss calculation for an input image (left), Grad-CAM (middle). Distance is represented using Viridis color-map in the right figure. Yellow is for the smallest distance and dark purple for the largest. In this case, the confounding region (in the lower left image region) that is wrongly found relevant is as far as it can be from the important region. Pixel intensity of Grad-CAM on the confounding region is exaggerated for presentation purposes.}
  \label{figure:chart_gradcam_contour_distance}
\end{figure}

We view the training images as instances made up of three parts: (1) the relevant regions, masked by $A_{obj}$, that are considered important for category classification; (2) the confounding regions, masked by annotation $A_{con}$, which are not correlated with any category but can trick the learner into learning that they are; and (3) the remaining image parts that are usually easily ignored by a learner as background image regions.  

Our explanation loss penalizes a learner based on the amount of wrong attention it gives to $A_{con}$, with due consideration of this wrong attention's distance from $A_{obj}$. For example, in Fig. \ref{figure:chart_gradcam_contour_distance}, a Grad-CAM explanation shows a model giving attention to a confounder located on the lower left corner of an input image. Distance of the attention to the confounder is illustrated with a Viridis color-map showing largest distances as dark purple. In this case, this would result in the highest penalty. As the model's (wrong) focus starts to get closer to $A_{obj}$, the explanation loss would decrease. We used Grad-CAM because it was found to be more sensitive to training label reshuffling and model parameter randomization \cite{adebayo2018sanity} than other saliency based explanations.

Equations \eqref{formula:proposed_explanation_loss} and \eqref{formula:proposed_final_loss} underpin how we propose to integrate explanation and classification losses. Algorithm \ref{alg:process} shows how this combined loss function is integrated into the overall XBL-D approach. Here, $G_{n}$ is the center of gravity of objects of interest in input images that are masked with $A_{obj}$, $expl_{\theta}(x_{n})$ is a Grad-CAM explanation of input $x_{n}$ to model $F$, and $A_{con}$ is the annotation of a  confounding region in $x_{n}$. A model's incorrect focus on a confounding region is detected using the intersection $I_{\theta}(x_{n})$ between $expl_{\theta}(x_{n})$ and $A_{con}$. The distance between a model's wrong attention to a confounding region and center of $A_{obj}$ or $G_{n}$ is then approximated by calculating average of the minimum and maximum euclidean distances, $d$, between points in $I_{\theta}(x_{n})$ and $G_{n}$. This gives us a measure of how far a model's incorrect attention is from the relevant image regions. In \eqref{formula:proposed_final_loss}, $L_{CE}$ represents the cross entropy loss and $\lambda{}\sum_{i=1}\theta_i$ is a weight ($\theta$) regularization term.

\begin{eqnarray}
\label{formula:proposed_explanation_loss}
    L_{expl} = \sum_{n}^N d(G_{n}, I_{\theta}(x_{n})) \\
\label{formula:proposed_final_loss}
    L =  \lambda_{1} L_{CE} + \lambda_{2} L_{expl} + \lambda{}\sum_{i=1}\theta_i
\end{eqnarray} 

\begin{algorithm}
    \caption{Distance-aware eXplanation Based Learning (XBL-D)}
    \label{alg:process}
    \textbf{Input}: confounded training dataset $\hat{X}$ and ground-truth category $Y$, feature annotation of object(s) of interest in $\hat{X}$: $A_{obj}$,  feature annotation of confounders in $\hat{X}$: $A_{con}$. \\
    \textbf{Parameters}: classification loss coefficient: $\lambda_{1}$, explanation loss coefficient: $\lambda_{2}$, regularization term: $\lambda$, network parameters: $\theta$\\
    \textbf{Output}: refined function ${F}$
    \begin{algorithmic}[1]
        \STATE $F$ $\leftarrow$ Fit function using $\hat{X}$ 
        \REPEAT
        \STATE $G$ $\leftarrow$ center of gravity of objects of interest in $A_{obj}$.
        \STATE $expl_{\theta}$ $\leftarrow$ saliency map explanations of $\hat{X}$ generated using Grad-CAM.
        \STATE $I_{\theta}$ $\leftarrow$ set of intersections between $expl_{\theta}$ and $A_{con}$
        \STATE $L_{expl}$ $\leftarrow$ explanation loss as average of the minimum and maximum euclidean distances between points in $I_{\theta}$ and $G$ 
        \STATE $L_{CE}$ $\leftarrow$ classification loss between $Y$ and $F(\hat{X})$
        \STATE Total loss, $L$ $\leftarrow$ $\lambda_{1}$ * $L_{CE}$ + $\lambda_{2}$ * $L_{expl}$ +  $\lambda{}\sum_{i=1}\theta_i$
        \STATE update $F$ using $L$
        \UNTIL{$L\leq\sigma$, where $\sigma$ is a tolerable total loss}
        \STATE \textbf{return} $F$
    \end{algorithmic}
\end{algorithm}

\subsection{Activation recall}

In addition to using AP, we propose Activation Recall (AR) to assess visual explanations, such as Grad-CAMs, generated by a trained model. AR measures how much of the relevant parts of test images are considered relevant by a model. This is presented in \ref{formula:activation_recall}, where (similarly to AP) $T_{r}$ is a threshold function that finds the (100-$r$) percentile and sets elements of $expl_\theta(x_{n})$ below this value to zero and the remaining elements to one.

\begin{eqnarray}
\label{formula:activation_recall}
    AR = \frac{1}{N} \sum_{n}^N\frac{ T_{r}(expl_{\theta}(x_{n})) * A_{obj_{n}}}{A_{obj_{n}}} 
\end{eqnarray} 

Instead of selecting a single threshold to assess generated explanations, we compute AP and AR at different thresholds to show impacts of choosing threshold on the evaluation metrics. This also gives us an insight into how explanation evaluation can be misleading without the full information, i.e thresholding.

\section{Experiments}
\label{section:methods}

In this section, we describe the  datasets, model architectures, and training details used in our experiments to evaluate the performance of XBL-D.

\subsection{Dataset}
\label{subsection:datasets}

\begin{figure}[t]
  \centerline{\includegraphics[width=1\linewidth]{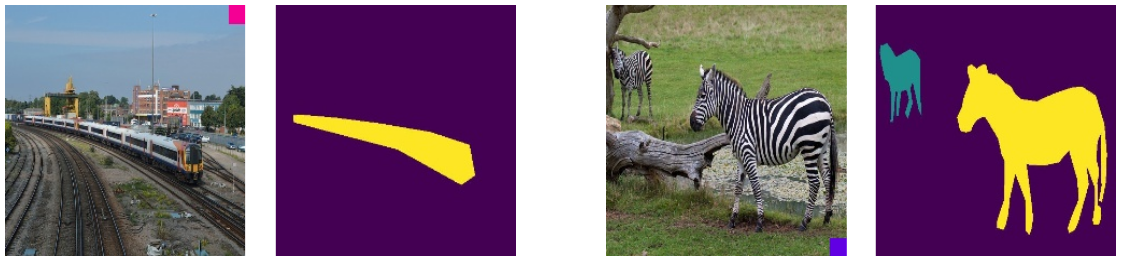}}
  \caption{Sample images from MS COCO with confounding regions added to random corners and their corresponding object masks}
  \label{figure:coco_sample_images}
\end{figure}

In order to validate performance of XBL-D, locations of the confounding regions needs to be known beforehand. For this reason, we used a publicly available decoyed dataset and created two new decoyed versions of existing datasets for our experiments:
\begin{enumerate}
    \item Decoy Fashion MNIST\footnote{We collected this dataset at \url{https://codeocean.com/capsule/7818629/tree/v1}}. This was created by Teso and Kersting \cite{teso2019explanatory}. 4x4 pixel confounders with random pixel intensities were added to random corners of images from the Fashion MNIST training dataset \cite{xiao2017fashion}. The 10,000 images from the test dataset were left clean.
    
    \item Decoy CIFAR-10\footnote{\url{https://osf.io/w5f7y/?view_only=abb7f5f55bfc48fb8c891838f699c0d3}}. We created this dataset by adding 4x4 pixel confounders with random pixel intensities to random corners of the training set of CIFAR-10 dataset. The CIFAR-10 dataset contains a training set of 50000 and test set of 10000 32x32 RGB images categorized into 10 classes. Similar to the Decoy Fashion MNIST, the test set of this dataset was also left clean for evaluation purposes.
    
    \item Decoyed subset of MS-COCO. We extracted a total of 2000 images for training and 600 image for testing, from the  \emph{Train} and \emph{Zebra} categories of the MS-COCO dataset \cite{lin2014microsoft}.  We then added 16x16 confounding pixels with random pixel intensities to random corners of the training images, which are of size 224x224. Images in the test set were left clean. We selected the \emph{Train} and \emph{Zebra} categories based on the low intersection of objects from both categories. Sample images from this dataset are shown in Fig. \ref{figure:coco_sample_images}. We refer to this dataset as Decoy MS-COCO$_{(2)}$.
\end{enumerate}

\subsection{Architecture selection and training}
\label{subsection:arch_selection}

We performed all of our experiments using Tensorflow and Keras\footnote{\url{https://www.tensorflow.org/api_docs/python/tf/keras}}. For all our datasets, we searched for the best model architectures and hyper-parameters using HyperBand algorithm \cite{li2017hyperband} in Keras tuner\footnote{\url{https://keras.io/keras_tuner/}}. We considered and optimized the hyper-parameters: number and size of convolutional layers, number of pooling layers, number and size of fully connected layers, and learning rate. A Convolutional Neural Network (CNN) with one convolutional layer containing 160 filters and two fully connected consecutive layers of sizes 992 and 800 nodes, and a learning rate = 1.158e-04 was found to perform best for the Decoy Fashion MNIST dataset. For the Decoy CIFAR-10, a CNN with two convolutional layers of filters 250 and 300 followed by one fully connected layer with 912 nodes, and a learning rate = 1.267e-04 was selected.  Similarly, we found that a CNN with four convolutional layers (containing 160, 352, 416, and 224 consecutive filters) each followed by a max-pooling layer, one fully connected layer of size 480 nodes, and a learning rate = 1.789e-05 performed best for the Decoy MS-COCO$_{(2)}$ dataset. 

To start with, the selected model architectures are fitted on the corresponding dataset using categorical cross-entropy loss and the Adam optimizer. We refer to the resulting models as \emph{Unrefined}. All the models are then refined using XBL-D. For the Decoy MS-COCO$_{(2)}$ dataset, we run 20 epochs of refinement where each epoch took an average of 15 minutes, while for each of the Decoy CIFAR-10 and Decoy Fashion MNIST datasets, we run 50 epochs of refinement each taking averages of 7 and 5 minutes, respectively. Model training was performed on a machine with NVIDIA RTX A5000 graphics card. 

Before starting the model refinement using XBL-D, we searched for optimal values of the coefficients of the categorical cross entropy loss ($\lambda_{1}$) and explanation loss ($\lambda_{2}$) using HyperBand in Keras tuner and we ended up with $\lambda_{1}=2.7$ and $\lambda_{2}=0.1$. We searched all hyper-parameters for each of the datasets separately. However, since $\lambda_{1}$ and $\lambda_{2}$ influence how XBL-D works, we decided to find one set that should work for the other domains for domain transferability purposes. Hence, the hyper-parameter search of $\lambda_{1}$ and $\lambda_{2}$ was performed on the most challenging task among the 3 datasets, which is the decoy MS-COCO$_{(2)}$ that contains large RGB images. 

\section{Results}
\label{section:results}

In this section, we present classification and explanation performance results of our proposed method and compare them against baseline methods.

\subsection{Classification}

Table \ref{table:fmnist_accuracy} presents classification accuracy performance of XBL-D and comparison against baseline methods. On the original test set of Fashion MNIST dataset, our proposed method achieves classification performance of 0.904 surpassing previous XBL methods \cite{friedrich2023typology}. The second best performing model was RRR with a classification accuracy of 0.894. None of the available baseline methods were implemented for our Decoy CIFAR-10 and Decoy MS-COCO$_{(2)}$. For this reason, we trained a model using the best performing method, RRR, on the decoyed CIFAR-10 and MS-COCO$_{(2)}$ datasets for comparison purposes. Again, compared to RRR and Unrefined models, XBL-D achieved superior classification performance on the original test sets of CIFAR-10 and MS-COCO$_{(2)}$  achieving accuracies of 0.843 and 0.938, respectively, as is summarized in Table \ref{table:fmnist_accuracy}.

\begin{table}[b]
\begin{center}
{\caption{Classification accuracy comparisons on original test images of Fashion MNIST, CIFAR-10, and MS-COCO$_{(2)}$.}\label{table:fmnist_accuracy}}
\begin{tabular}{p{0.1\columnwidth} p{0.2\columnwidth}p{0.2\columnwidth}p{0.3\columnwidth}}
\hline
\textbf{Method}           & \textbf{Decoy Fashion MNIST}  & \textbf{Decoy CIFAR-10} & \textbf{Decoy MS-COCO$_{(2)}$}       \\ 
\hline
Unrefined        & 0.862  & 0.789  & 0.845        \\ 
XBL-D & \textbf{0.904} & \textbf{0.843} & \textbf{0.938} \\
RRR              & 0.894  & 0.810   & 0.853    \\
RRR-G            & 0.786  & -  &  - \\
RBR              & 0.876   &  - &-       \\
CDEP             & 0.767   &  - &-       \\
HINT             & 0.582   & - &-        \\
CE                 & 0.858   & - &-       \\
\hline
\end{tabular}
\end{center}
\end{table}

\begin{table}[b]
\begin{center}
{\caption{Summary evaluations of explanations generated for the original Fashion MNIST, MS-COCO and CIFAR-10 test datasets.}\label{table:summary_explanation_evaluation}}
\begin{tabular}{p{0.06\columnwidth}p{0.11\columnwidth}p{0.2\columnwidth}p{0.2\columnwidth}p{0.2\columnwidth}} 
\hline
\textbf{Metric}              & \textbf{Method}    & \textbf{Decoy Fashion MNIST} & \textbf{Decoy CIFAR-10} & \textbf{Decoy MS-COCO$_{(2)}$}   \\
\hline
\multirow{3}{*}{AR} & Unrefined & 0.280  & 0.419 & 0.500   \\
                    & XBL-D     & \textbf{0.557}  & \textbf{0.516} & \textbf{0.860}  \\
                    & RRR       & 0.335  & 0.432 & 0.841 \\
\hline
\multirow{3}{*}{AP} & Unrefined & 0.318 & 0.168 & 0.609  \\
                    & XBL-D     & \textbf{0.663} &   \textbf{0.342}  & 0.698     \\
                    & RRR       & 0.425  &  0.181 & \textbf{0.761}
                    \\
                    \hline
\end{tabular}
\end{center}
\end{table}


\subsection{Explanation performance}

\begin{figure}[t]
  \centerline{\includegraphics[width=\linewidth]{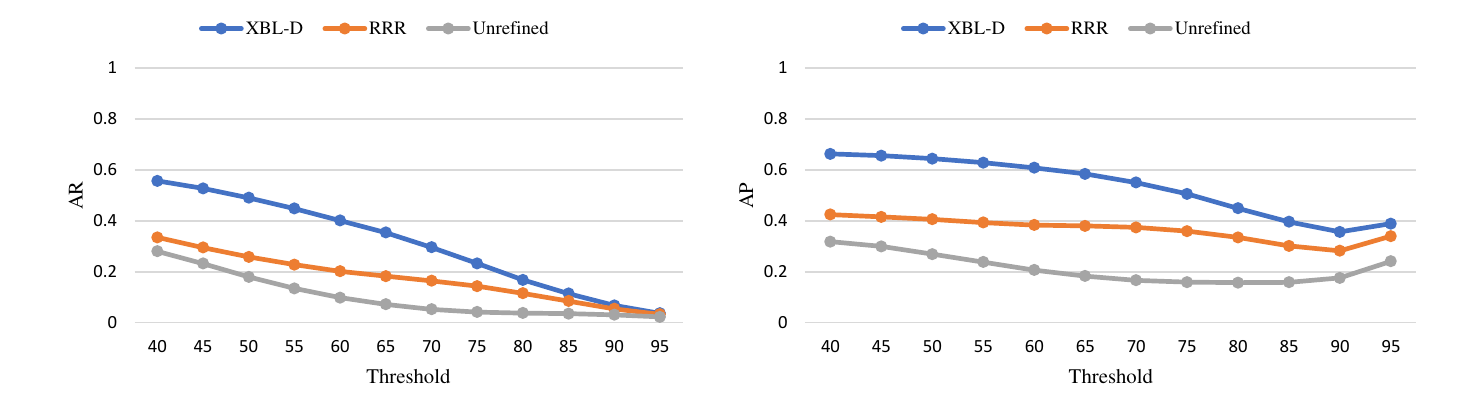}}
  \caption{AR and AP evaluations of explanations generated for a clean Fashion MNIST test dataset using a model trained on the Decoy Fashion MNIST. The evaluations are performed at threshold values ranging from 40\% to 95\% with step size = 5}
  \label{figure:chart_ar_ap_fmnist}
\end{figure}

\begin{figure}[t]
  \centerline{\includegraphics[width=\linewidth]{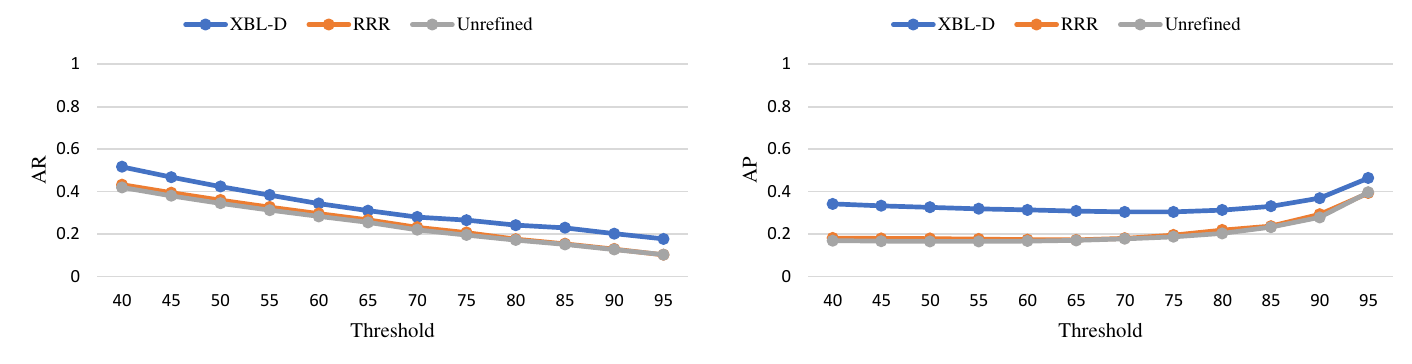}}
  \caption{AR and AP evaluations of explanations generated for a clean CIFAR-10 test dataset using a model trained on the Decoy CIFAR-10. The evaluations are performed at threshold values ranging from 40\% to 95\% with step size = 5}
  \label{figure:chart_ar_ap_cifar}
\end{figure}

\begin{figure}[t]
  \centerline{\includegraphics[width=\linewidth]{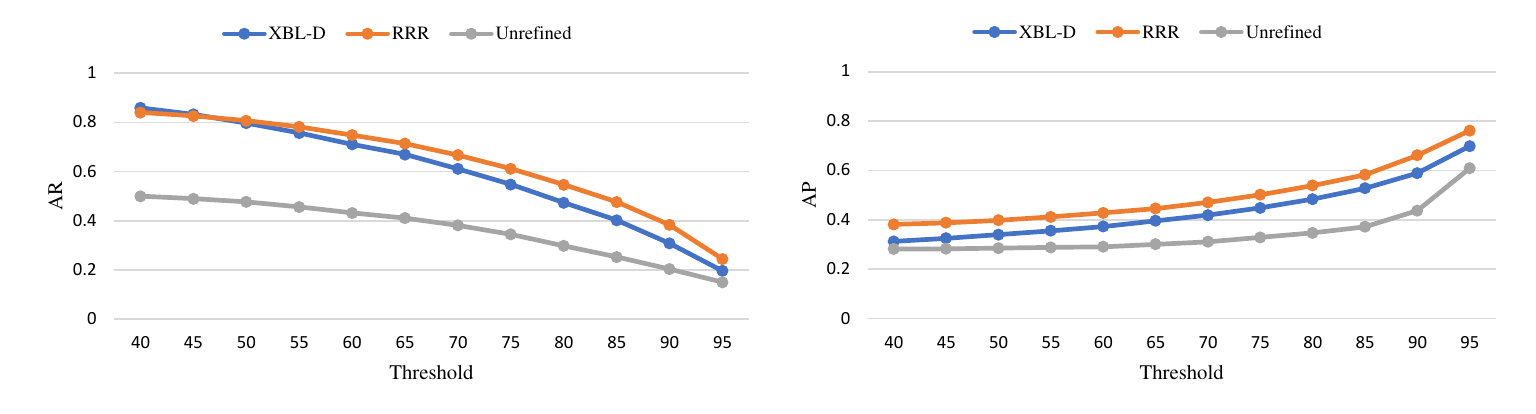}}
  \caption{AR and AP evaluations of explanations generated for a clean MS-COCO test dataset using a model trained on the Decoy MS COCO$_{(2)}$. The evaluations are performed at threshold values ranging from 40\% to 95\% with step size = 5}
  \label{figure:chart_ar_ap_coco}
\end{figure}

While AR and AP evaluations, on the original test sets of the Fashion MNIST, MS-COCO$_{(2)}$ and CIFAR-10, across thresholds ranging from 40\% to 95\% with step size = 5 are presented in Figures \ref{figure:chart_ar_ap_fmnist}, \ref{figure:chart_ar_ap_cifar}, and \ref{figure:chart_ar_ap_coco}, Table \ref{table:summary_explanation_evaluation} presents a summary of evaluations of explanations.

\subsubsection{Fashion MNIST} 

Our proposed method scores higher than both RRR and Unrefined models using both metrics. At threshold = 40\%, XBL-D scored highest values of AR = 0.557 and AP = 0.663 (Table \ref{table:summary_explanation_evaluation}). Given that higher threshold means considering smaller areas of Grad-CAM, AR values decrease with increasing threshold (see Fig. \ref{figure:chart_ar_ap_fmnist}). However, even though AP seemed to decrease with increasing threshold values, it starts to increase at threshold above 90\% (we accredit this to the Gray-Scale nature of the decoy Fashion MNIST dataset).

\subsubsection{CIFAR-10} 

Similar to the Fashion MNIST, our proposed method performs higher than both RRR and Unrefined models using both metrics. At threshold = 40\%, XBL-D scored highest values of AR = 0.516 (Table \ref{table:summary_explanation_evaluation}) and at threshold = 95\%, XBL-D scored AP = 0.342, outperforming both methods. While AR naturally decreases with increasing threshold, AP increases given the RGB nature of CIFAR-10.


\begin{figure*}[h!]
  \centerline{\includegraphics[width=0.7\linewidth]{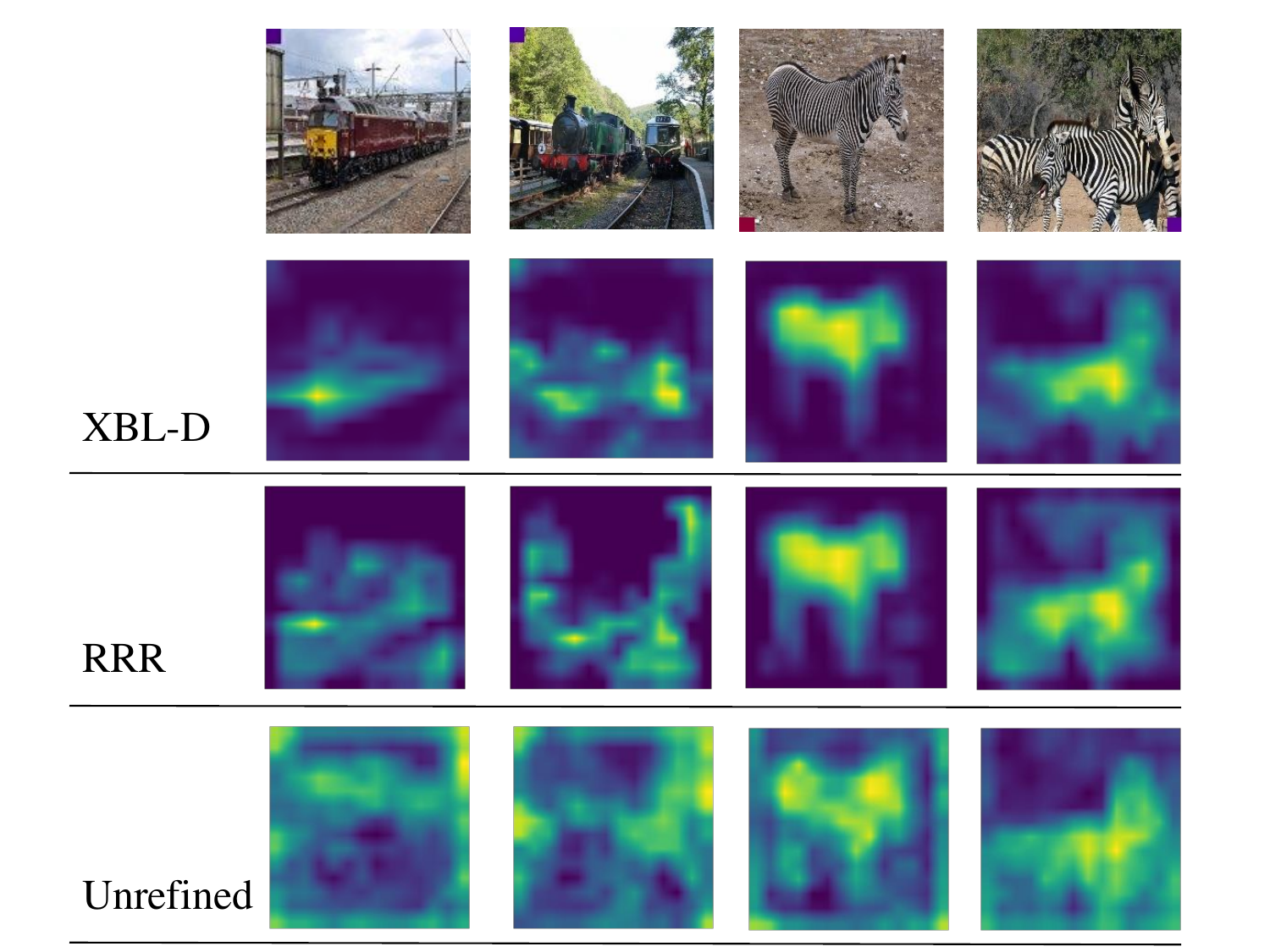}}
  \caption{Sample Grad-CAM Outputs. Original size of all Grad-CAM images was 14x14; They are up-sampled to 224x224 for easier comparison against input images.}
  \label{figure:chartsample_gcams}
\end{figure*}

\subsubsection{MS-COCO$_{(2)}$} 

Our method scored better AR at lower thresholds and performed comparable to RRR at other thresholds (at threshold = 40\%, XBL-D scored AR = 0.860, Table \ref{table:summary_explanation_evaluation}). Similar to the other datasets, we also found that low threshold values led to higher AR values (see Fig. \ref{figure:chart_ar_ap_coco}). However, unlike the models trained on the Fashion MNIST dataset but similar to the CIFAR-10, AP values increase with increasing threshold (at threshold=95\%, RRR scored highest AP of 0.761, Table \ref{table:summary_explanation_evaluation}). We accredit this to the RGB nature of the MS-COCO$_{(2)}$ dataset.


Sample Grad-CAM outputs of input images from both categories are displayed in Fig. \ref{figure:chartsample_gcams}. We show sample explanation outputs for the MS-COCO$_{(2)}$ images because their high resolution makes them well suited for presentation. While the clean test sets were used in computing AR and AP explanation evaluations, sample of the decoyed images from training set of the MS-COCO$_{(2)}$ are shown in Fig. \ref{figure:chartsample_gcams} to demonstrate the ability of XBL-D in avoiding confounding regions and to compare it against RRR and the Unrefined model. As is displayed in the sample outputs, our proposed method was able to produce accurate explanations that focus on relevant parts of objects in input images and successfully ignores confounders.

\section{Discussion}
\label{section:discussion}


In addition to explaining a model's classification output, XBL facilitates a more transparent machine learning process by providing a rich user interaction mechanism. As opposed to the traditional interactive machine learning that is usually performed through instance category labeling, a user would be able to get involved at a deeper level by interacting with model explanations in the machine learning process. In XBL, a user would be able to teach a learner model by observing and commenting on the reasoning (i.e correcting model explanations) behind its predictions. This kind of user engagement has the potential to circumvent the \emph{black-box} public image of deep learning models since it aims to build a rapport with users by providing a transparent way of interaction with an opportunity to refine the models.

When compared against baseline methods, XBL-D achieved superior performance in classifying all three datasets. We believe this is because it unlearns confounding regions, which were wrongly found relevant by a model, based on their locations and distances from the user annotated relevant regions. As shown in the sample outputs in Fig. \ref{figure:chartsample_gcams}, a model's focus, shown with visual explanations is not strictly bounded and, however good it is, there is always a good chance it might exceed boundaries of relevant region(s). Based on this fact, XBL-D instructs a learner that it is not only acceptable to focus on the user annotated parts but also around it as long as it keeps a distance from the confounding region. Had the explanation loss been based on intersection of generated explanations with the confounding regions, it would penalize the model whenever it focuses on the confounders without consideration for the confounders' locations. 

In addition to XBL-D, we observe that the Unrefined model performed better than most of the other XBL models in classifying the decoy Fashion MNIST. We attribute this to the accuracy-interpretability trade-off in deep learning. Although the existence of this trade-off is debated \cite{rudin2019stop,dziugaite2020enforcing}, deep learning models that are refined with an explanation based learning could lose performance if the refinement is not performed using a fitting approach such as our proposed method, XBL-D.

We also proposed an interpretability metric, Activation Recall (AR). AR measures how much of the user annotated relevant image regions were actually considered relevant by a trained model. It circumvents a possible over-confidence that may result from mainly focusing on explanations (saliency maps in this case) during explanation evaluation. By redirecting the focus from explanations to ground-truth annotations, AR provides a reliable metric for explanation evaluation. We recommend AR should be used in conjunction with AP for a reliable assessment of model explanations.

Objective evaluations of generated explanations of test images of employed datasets across different thresholds also show that XBL-D performs better than RRR and Unrefined models in generating accurate explanations. Threshold selection is important in computing AR and AP of generated explanations. In all the datasets, we observed that low threshold values lead to higher AR while the opposite is true for AP. This is because parts of Grad-CAM considered for AR calculation increase with decreasing threshold. We also note that the Gray-Scale nature of Fashion MNIST affects AP values and it plummets with increasing threshold, but recovers after threshold = 90\%. 

In addition to performing better at objective evaluations, XBL-D also outputs visually accurate saliency maps compared to the RRR and Unrefined models as can be seen in Fig. \ref{figure:chartsample_gcams}. We were able to observe that XBL-D is better than RRR and the Unrefined models at localizing objects of interest in input images.

\section{Conclusion}
\label{section:conclusion}

In this paper we proposed and demonstrated superior performance of XBL-D, a distance-aware explanation loss for XBL loss function augmentation. This introduces a new direction for XBL research: the consideration of the distance of a model's wrong attention from relevant regions. XBL-D was able to achieve superior classification and interpretability performance compared to baseline methods on three different datasets. This assures that our proposed method generalizes across different datasets.

\section*{Acknowledgment} This publication has emanated from research conducted with the financial support of Science Foundation Ireland under Grant number 18/CRT/6183. For the purpose of Open Access, the author has applied a CC BY public copyright licence to any Author Accepted Manuscript version arising from this submission.

\bibliographystyle{IEEEtran}
\bibliography{splncs04}

\end{document}